\documentclass[conference]{IEEEtran}
\ifCLASSINFOpdf
 
\else

\fi

\usepackage[utf8]{inputenc}
\usepackage{xcolor}
\usepackage{cite}
\usepackage{amsmath,amssymb,amsfonts}
\usepackage{algorithmic}
\usepackage{graphicx}
\usepackage{textcomp}
\usepackage{xcolor}
\usepackage{float}
\usepackage{hyperref}
\usepackage{flushend}
\usepackage{comment}
\usepackage{subfigure}

\usepackage{booktabs}
\usepackage{tabularx}
\newcolumntype{Y}{>{\centering\arraybackslash}X}

\usepackage{url}
\usepackage{pifont}
\usepackage{tabulary}

\begin{document}

\title{A Fast Fully Octave Convolutional Neural Network for Document Image Segmentation}

\author{\IEEEauthorblockN{
Ricardo Batista das Neves Junior\,$^{1}$, 
Luiz Felipe Verçosa\,$^{1}$,
David Macêdo\,$^{2,3}$,\\
Byron Leite Dantas Bezerra\,$^{1}$, and
Cleber Zanchettin\,$^{2,4}$
}
\IEEEauthorblockA{\,
$^{1}$Escola Politécnica de Pernambuco, Universidade de Pernambuco, Recife, Brasil\\
$^{2}$Centro de Inform\'atica, Universidade Federal de Pernambuco, Recife, Brasil\\
$^{3}$Montreal Institute for Learning Algorithms, University of Montreal, Quebec, Canada\\
$^{4}$Department of Chemical and Biological Engineering, Northwestern University, Evanston, United States of America\\
Emails: \{rbnj, lfvv\}@ecomp.poli.br, dlm@cin.ufpe.br, byron.leite@upe.br, cz@cin.ufpe.br}
}

\markboth{Journal of \LaTeX\ Class Files,~Vol.~14, No.~8, August~2015}%
{Shell \MakeLowercase{\textit{et al.}}: Bare Demo of IEEEtran.cls for IEEE Journals}

\maketitle
\begin{abstract} 
The Know Your Customer (KYC) and Anti Money Laundering (AML) are worldwide practices to online customer identification based on personal identification documents, similarity and liveness checking, and proof of address. To answer the basic regulation question: are you whom you say you are? The customer needs to upload valid identification documents (ID). This task imposes some computational challenges since these documents are diverse, may present different and complex backgrounds, some occlusion, partial rotation, poor quality, or damage. Advanced text and document segmentation algorithms were used to process the ID images. In this context, we investigated a method based on U-Net to detect the document edges and text regions in ID images. Besides the promising results on image segmentation, the U-Net based approach is computationally expensive for a real application, since the image segmentation is a customer device task. We propose a model optimization based on Octave Convolutions to qualify the method to situations where storage, processing, and time resources are limited, such as in mobile and robotic applications. We conducted the evaluation experiments in two new datasets \textit{CDPhotoDataset} and \textit{DTDDataset}, which are composed of real ID images of Brazilian documents. Our results showed that the proposed models are efficient to document segmentation tasks and portable.
\end{abstract}


\IEEEpeerreviewmaketitle

\section{Introduction}

The segmentation task for automatic text detection and document region identification is essential on online user or customer identification for companies, public entities, among others. This is also a regulated and mandatory task based on Know Your Customer (KYC) \cite{mullins2014know} as part of the Anti Money Laundering (AML) \cite{sharman2008power}, which refers to a set of laws, regulations, and procedures intended to prevent criminals from disguising illegally obtained funds as legitimate income. 

One of the first steps in the KYC process to verifying the identity of the clients and assessing their suitability is the uploading of the client identification documents (ID). This is a challenging task that has increased since now the customer is allowed to upload their own photos instead of attending to a physical office for document digitalization.
This flexibility increased access to services and products but eliminated the digitalization standards and controls. The pictures now come from all sorts of devices such as smartphones, digital cameras, scanners, and others. 

Another obstacle is that companies usually accept different kinds of identification documents, such as ID, driver license, and professional license, that are now captured without training and standard procedures. The users are capturing more than one document per photo, with different and complex backgrounds, eventually occlusion, partial rotation, no focus, poor quality, or even damaged. In this context, automatic image segmentation is a crucial preprocessing step for the final tasks as document classification, text recognition, face recognition, signature verification, and user identification \cite{mello2012}. 

Algorithms based in Convolutional Neural Networks (CNN) have been successfully applied to segment images of identification documents \cite{forczmanski2019segmentation} \cite{oliveira2018dhsegment} \cite{renton2017handwritten}. Nevertheless, standard CNNs require a large number of images for training that might not be available. In this context, U-Net \cite{ronneberger2015u}, which is a Fully Convolutional Neural Network, has been successfully applied to data segmentation tasks and uses data augmentation to overcome dataset size limitations. 

U-Net consists of a contracting and an expanding path in a u-shape form. The contracting party can capture context information, while the expanding part allows for the precise location of parts of the input image.

However, the large number of parameters and resources needed for previous approaches may not be viable on mobile devices, where computational resources such as memory and processing capacity are insufficient to the point of compromising the quality and performance of real-time applications. Besides the promising results and reduced size of the standard U-Net in comparison to other systems based on CNNs, even this approach is still unable to be applied to real applications on portable or embedded devices.

On the other hand, Octave Convolutions \cite{chen2019drop} has been able to reduce computational costs while obtaining satisfactory results \cite{fan2019accurate}. The Octave Convolutions uses a strategy of multi-frequency feature representation. It decomposes the image in its low and high spatial frequency parts, storing low-frequency maps in low-resolution tensors eliminating unnecessary redundancy. Therefore, according to studies presented in the literature \cite{lundgren2019octshufflemlt}, it is expected that image segmentation algorithms based on Octave Convolutional approach, obtain similar accuracy to current CNN models using traditional convolutions, despite being able to get a lower computational cost model.

In this paper, we propose a new approach to image segmentation of ID documents considering the segmentation performance in real documents and the applicability on portable devices. Our main contributions are:
\begin{enumerate}
    \item A Fully Octave Convolutional Neural Network (FOCNN or FOCN) based on the U-Net architecture here called OctHU-PageScan. In this approach, we reduce the redundancy of feature maps and generate a lighter model, which can be applied in mobile and robotic applications. The experiments with OctHU-PageScan show that its processing time is up to 75.13\% and 73.79\% faster than the U-Net on the proposed datasets \textit{CDPhotoDataset}, and \textit{DTDDataset} respectively, using 93.49\% less disk storage space when compared to the U-Net and having 93.52\% less trainable parameters.
    \item Two new datasets to document image segmentation:
    \begin{enumerate}
        \item \textit{CDPhotoDataset}: A dataset composed of 20,000 images of Brazilian documents. While other literature datasets (e.g, Smartdoc dataset \cite{burie2015icdar2015}) presents documents on a uniform background, our images containg documents with different complex real-world backgrounds. 
        \item \textit{DTDDataset}: A dataset of 10,000 document images (without a background) that can be used for text detection task.
    \end{enumerate}
    
\end{enumerate}

The next sections of this paper are organized as follows: In Section II, we shortly present previous works tackled with similar problems. In Section III, we present our proposed methods. Section IV describes the experiment's setup, and Section V presents the results and discussion. In Section VI, presents our conclusions and possible future research.

\section{Background} \label{background}

This work investigates Deep Learning techniques for image segmentation in  ID documents. Among the leading models in the literature for this purpose, approaches based on CNNs have received much attention due to their excellent generalization capacity in training image datasets applied to the most diverse tasks, such as segmentation, classification, among others.

Forczmanski et al. \cite{forczmanski2019segmentation} proposed the segmentation of text from paper documents using a CNN based approach. The authors used a CNN model to object detection based on the confidence level. The Forczmanski approach is general and aims to find logos, stamps, texts, signatures, and tables in an image document.

Oliveira et al. \cite{oliveira2018dhsegment} developed a generic CNN based approach called dhSegment to perform segmentation in historical documents. The authors created a pixel-wise predictor coupled with task-dependent post-processing blocks. The author's goal is to find text in handwritten historical documents.

Renton et al. \cite{renton2017handwritten} developed a technique based on a variant of Fully Convolutional Networks (FCN) that relies on dilated convolutions. The method uses an x-height labeling of the text line that improves the text line recognition. 

Some works in the literature used CNN for segmentation in images of commercial documents. Melo et al. (2018) \cite{melo2018fully} proposed an FCN for signature segmentation on commercial documents such as checks, contracts, and IDs, using the signatures extracted from the MCYT dataset \cite{fierrez2004off}. Silva et al. (2019) \cite{paloma2019} presented an optimized FCN architecture based on U-Net for the same problem. In this case, the architecture proposed by the authors has fewer parameters and obtained better results than the baseline model proposed in \cite{ melo2018fully}.

One drawback of CNN approaches is the time necessary for training and inference due to the number of parameters, layers, and setup required. Even the previous model \cite{paloma2019} also uses storage, memory, processing power, and inference time incompatible in case of real-time applications running in mobile devices.

A recent technique called Octave Convolution was presented in \cite{chen2019drop}, aiming to reduce both memory and computational cost while assuring the same results when compared to technologies such as Convolutional Neural Networks (CNN) with traditional convolutions. The method factorizes the feature maps by frequency and performs an operation to store and process the feature maps acting as a plug-and-play convolutional unit. This approach is capable of extracting high and low-frequency information, increasing the capacity of obtaining specific details with sudden changes between foreground and background. 

\begin{figure}
    \centering
    \includegraphics[width=\columnwidth,keepaspectratio]{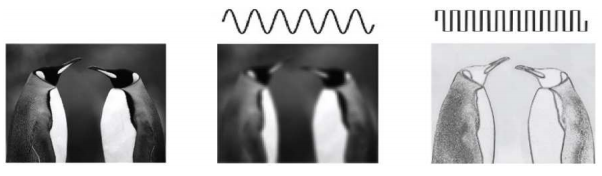}
    \caption{Image reproduced and adapted from \cite{chen2019drop}. The image on the left illustrates the original image, the image on the center illustrates the original image after separating only the low-frequency signal and the image on the right shows the original image after separating only the high-frequency signal.}\label{octave_reproduced}
\end{figure}

Some related works have used Octave Convolution in the area of document segmentation, but not of identification documents. Figure \ref{octave_reproduced} shows how the separation of low and high-frequency signals works. In the low-frequency signal, the image structure changes smoothly, while in the high-frequency signal, it describes the fine details that change quickly.

A few approaches were found by using Octave Convolution since it is a recent technique. Lundgren et al. \cite{lundgren2019octshufflemlt} proposed an improvement in Octave Neural Network for detecting multilingual scene texts. Some of the improvements made are related to the usage of FPN Object Detector using ShuffleNets \cite{zhang2018shufflenet}. 

Fan et al. \cite{fan2019accurate} tackled the problem of segmentation of retinal vessel using encoder-decoder based on Octave Convolution. They also included a new operation called octave transposed convolution. They obtained better or comparable performance results when comparing to their CNN baseline.

\section{OctHU-PageScan}

In this Section, we present OctHU-PageScan, a Fully Octave Convolutional Network for image segmentation, foreground document edge, and text detection. The model receives as input a grayscale image of shape 512x512x1 and for the training a ground truth (GT), of the same shape, in black and white representation. The GT Image represents the interest regions in white and the background in black.

Table \ref{networkArchitetureTable} shows the architecture of the proposed network. The proposed approach is split into two parts: the encoding, which allows the reduction of image shape by extracting its main features, and the decoding part, which uses the extracted features to produce an output image with the interest regions highlighted. 

The encoding process applies two consecutive octave convolutions, followed by a rectified linear unit (ReLU) and max-pooling function. For each downsampling layer, the FOCN doubles the number of features map.

The decoding process is composed of upsampling to double image size, followed by 2x2 convolutions (up-convolutions), reducing by half the number of feature maps. Next, it is performed the concatenation with the respective layers of the encoding process to compose the characteristics of the decode. Next, two-octave convolutions 2x2 are applied, followed by the ReLU function. Finally, an octave convolution of $alpha = 0$ is used to map the components of the feature vector. The source code of our proposed system is available at \url{https://github.com/ricardobnjunior/OctHU-PageScan}.

\begin{table}
\scriptsize
\renewcommand{\arraystretch}{1.5}
\centering
\caption{Proposed model architecture. Total Parameters: 1,963,794}
\begin{tabular}{l|l|l|l|l}
\toprule
Layer & \begin{tabular}[c]{@{}l@{}}Features map\\ (output shape)\end{tabular} & Activation & Connected to & Alpha \\ \hline
Input & 1 x 512 x 512 & \textbf{} &  &  \\ \hline
oct\_conv\_1 & 16 x 512 x 512 & ReLU & Input & 0.5 \\ \hline
oct\_conv\_1 & 16 x 512 x 512 & ReLU & oct\_conv\_1 & 0.5 \\ \hline
Max\_Pool\_1 & 16 x 256 x 256 &  & oct\_conv\_1 &  \\ \hline
oct\_conv\_2 & 32 x 256 x 256 & ReLU & Max\_Pool\_1 & 0.5 \\ \hline
oct\_conv\_2 & 32 x 256 x 256 & ReLU & oct\_conv\_2 & 0.5 \\ \hline
Max\_Pool\_2 & 32 x 128 x 128 &  & oct\_conv\_2 &  \\ \hline
oct\_conv\_3 & 64 x 128 x 128 & ReLU & Max\_Pool\_2 & 0.5 \\ \hline
oct\_conv\_3 & 64 x 128 x 128 & ReLU & oct\_conv\_3 & 0.5 \\ \hline
Max\_Pool\_3 & 64 x 64 x 64 &  & oct\_conv\_3 &  \\ \hline
oct\_conv\_4 & 128 x 64 x 64 & ReLU & Max\_Pool\_3 & 0.5 \\ \hline
oct\_conv\_4 & 128 x 64 x 64 & ReLU & oct\_conv\_4 & 0.5 \\ \hline
Max\_Pool\_4 & 128 x 32 x 32 &  & oct\_conv\_4 &  \\ \hline
oct\_conv\_5 & 256 x 32 x 32 & ReLU & Max\_Pool\_4 & 0.5 \\ \hline
oct\_conv\_5 & 256 x 32 x 32 & ReLU & oct\_conv\_5 & 0.5 \\ \hline
upsize\_1 & 256 x 64 x 64 &  & oct\_conv\_5 &  \\ \hline
concatenate\_1 &  &  & upsize\_1, oct\_conv\_4 &  \\ \hline
oct\_conv\_6 & 128 x 64 x 64 & ReLU & concatenate\_1 & 0.5 \\ \hline
oct\_conv\_6 & 128 x 64 x 64 & ReLU & oct\_conv\_6 & 0.5 \\ \hline
upsize\_2 & 128 x 128 x 128 &  & oct\_conv\_6 &  \\ \hline
concatenate\_2 &  &  & upsize\_2, oct\_conv\_3 &  \\ \hline
oct\_conv\_7 & 64 x 128 x 128 & ReLU & concatenate\_2 & 0.5 \\ \hline
oct\_conv\_7 & 64 x 128 x 128 & ReLU & oct\_conv\_7 & 0.5 \\ \hline
upsize\_3 & 64 x 256 x 256 &  & oct\_conv\_7 &  \\ \hline
concatenate\_3 &  &  & upsize\_3, oct\_conv\_2 &  \\ \hline
oct\_conv\_8 & 32 x 256 x 256 & ReLU & concatenate\_2 & 0.5 \\ \hline
oct\_conv\_8 & 32 x 256 x 256 & ReLU & oct\_conv\_8 & 0.5 \\ \hline
upsize\_4 & 32 x 512 x 512 &  & oct\_conv\_8 &  \\ \hline
concatenate\_4 &  &  & upsize\_4, oct\_conv\_1 &  \\ \hline
oct\_conv\_9 & 16 x 512 x 512 & ReLU & concatenate\_4 & 0.5 \\ \hline
oct\_conv\_9 & 16 x 512 x 512 & ReLU & oct\_conv\_9 & 0.5 \\ \hline
oct\_conv\_10 & 1 x 512 x 512 & Sigmoid & oct\_conv\_9 & 0.0 \\
\bottomrule
\end{tabular}
\label{networkArchitetureTable}
\end{table}

\section{Materials and Methods}

In order to evaluate the proposed model for document segmentation, we considered two different tasks. The first one is the document boundary detection taking into account it may appear in any region of a sample image. Moreover, we considered a sample image might have several documents in any position and orientation (see Figure  \ref{datasetExample}).

The second segmentation task we have considered is the text segmentation. As a real scenario, we don't have any assumptions regarding the text size, position and orientation in documents. Therefore, in this section, we introduce the two new evaluation datasets built for investigating both problems and present the experimental setup. 

\subsection{Datasets}

The segmentation task consisted of identifying text regions in documents such as personal identification documents, driver licenses, and professional licenses. To create the labeled dataset, we used the VGG Image Annotator (VIA)\footnote{http://www.robots.ox.ac.uk/~vgg/software/via/}, free software that allows the creation of polygons to identify image regions such as text regions or document boundary in photos or scanned documents. After creating those regions, we obtained, through VIA, a JSON format file containing the corresponding pixel areas of the created polygons. These regions were used as ground truth.

\subsubsection{CDPhotoDataset}

The CDPhotoDataset is composed of images of Brazilian identification documents, in different backgrounds of real-world environments, with non-uniform standards, consisting of different colors, textures, lighting conditions, and different image resolutions. In addition, the documents in this dataset were acquired from different smartphone cameras in different lighting environments, as well as from multiple scanner devices.

\begin{figure}
    \centering
    \includegraphics[width=\columnwidth,keepaspectratio]{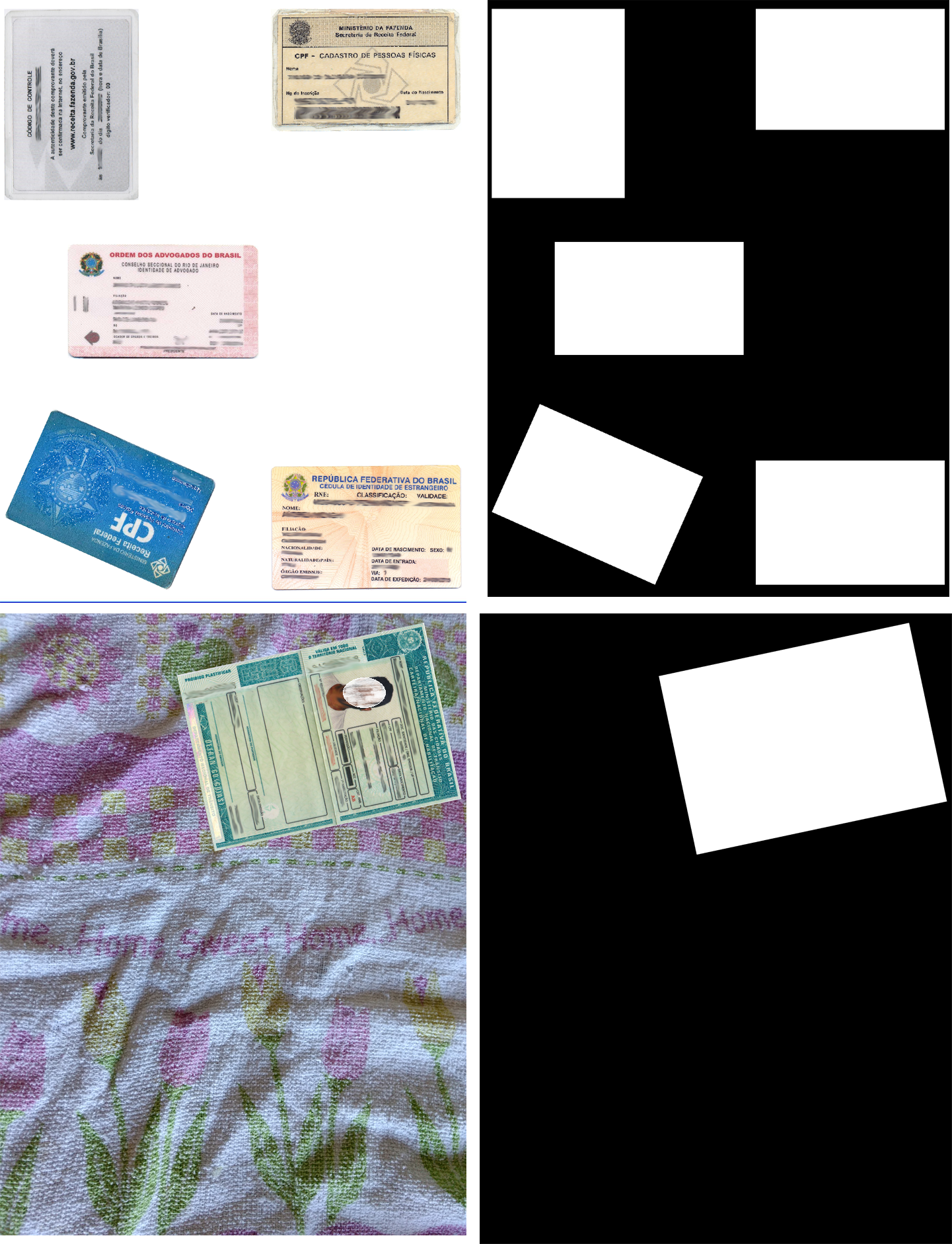}
    \caption{Example of two images present in the CDPhotoDataset. In the left, a synthetic image, in the right the respective mask of the target document (the ground truth).}\label{datasetExample}
\end{figure}

The CDPhotoDataset has a total of 20,000 images of documents with the respective document mask as ground truth. In order to have a representative dataset, different background types, resolutions, documents, and lighting conditions were applied to simulate the complexity and diversity of the real-world scenarios. Also, each identification document was mixed with some image background with a random perspective and noise. Moreover, a single image can be composed from one to five different types of documents, as sampled in Figure \ref{datasetExample}).

To compose the CDPhotoDataset, we used 273 variations of backgrounds (being 213 camera backgrounds and 60 scanner backgrounds), 40 types of distortions in perspective, and other aspects of the document. As a consequence of this procedure, the images included in the CDPhotoDataset are quite similar to photos of documents captured from the real-world. Figure \ref{datasetExample} shows two examples of these images.

\subsubsection{DTDDataset} \label{desc_DTDDataset}

DTDDataset is composed of Brazilian ID documents divided into eight classes: front and back faces of National Driver's License (CNH), CNH front face, CNH back face, Natural Persons Register (CPF) front face, CPF back face, General Registration (RG) front face, RG back face, and RG front and back faces.

\begin{figure}
    \centering
    \includegraphics[width=\columnwidth,keepaspectratio]{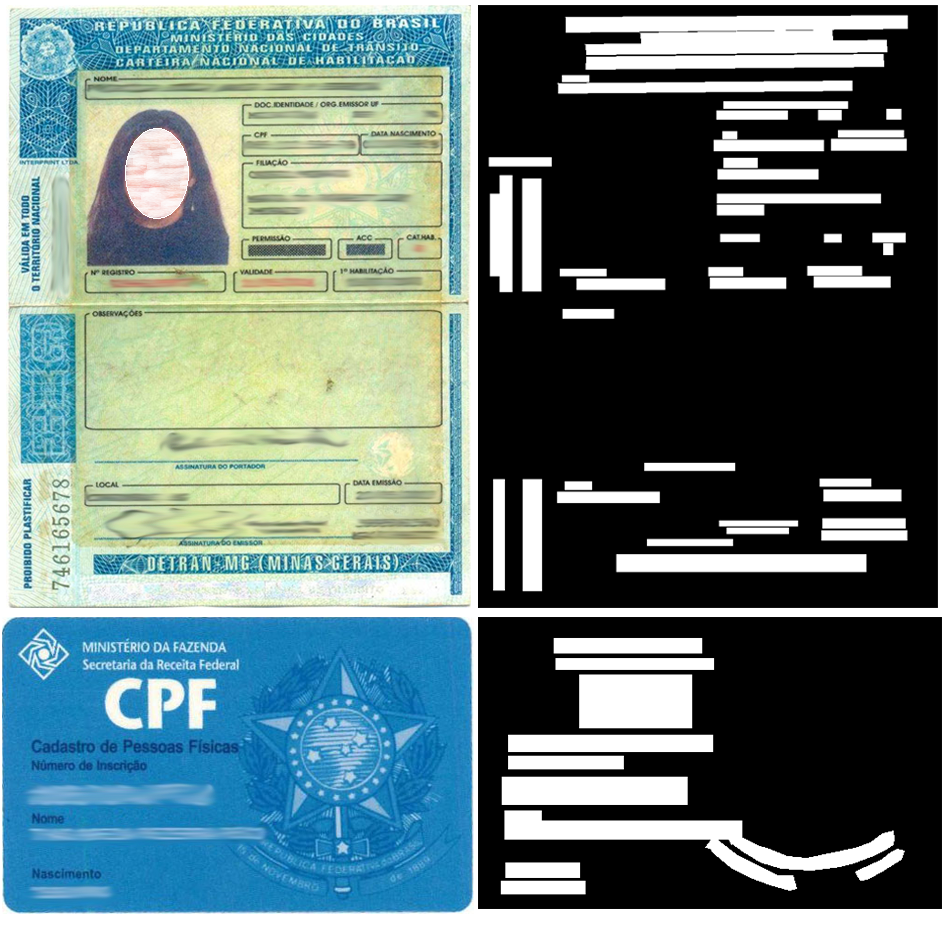}
    \caption{Example of two images present in the DTDDataset. In the left, a document image, in the right the respective mask (the ground truth).}\label{DocumentTextDetectionDatasetExample}
\end{figure}

The dataset is composed of the original image (document photo) and ground truth (mask delimiting the text region), as shown in Figure \ref{DocumentTextDetectionDatasetExample}. Initially, the dataset consisted of 100 documents from each class, yielding a total of 800 documents. We divided 130 documents to the testing set and applied a Data Augmentation in the rest of the dataset (training set).

The applied Data Augmentation has generated 10,000 new images for training, with different image modifications such as noise, brightness, rotation, distortion, zoom, occlusion, resizing, and flip.

\subsection{Training  Strategy}

It is worth to mention we have a separate training and evaluation analysis independent for each dataset since they correspond to different tasks in a full document segmentation and text recognition pipeline. In any case, the target dataset is divided into training and testing sets. 

In the case of the CDPhotoDataset, we randomly split it into 75\% for training and 25\% for testing. While DTDDataset has been organized as previously described (Section \ref{desc_DTDDataset}).

For training the models, we used the Adam Optimizer \cite{kingma2014adam}. To verify the similarity between the ground truth (GT) and the model output, we used the Jaccard Similarity Coefficient (JCS) \cite{hamers1989similarity}, presented in equation \ref{jaccard}:

\begin{equation}
J(A,B) = \frac{|A \cap B|}{|A \cup B|}
\label{jaccard}
\end{equation} where $A$ is the image inferred by the model and $B$ is the ground truth, taking into account that both, the image generated by the network output and the ground truth, are represented by binary values, where zero represents the black pixel, and one the white pixels.

The learning rate used was 0.0001, and the size of the mini-batches is equal to 4. The training lasted 10,000 epochs. In the whole process, we used a computer with the following settings: Intel Core i7 8700 processor, 16GB 2400Mhz RAM, and 6GB NVIDIA GTX 1060.

\section{Results and Discussion}

Table \ref{tabela_result_metr_jcs} shows the achieved results considering the models using JCS Rate. Our proposal decreased 4.6 percentual points in accuracy on \textit{CDPhotoDataset} and 16.51 points in the \textit{DTDDataset}. 

On the other hand, Table \ref{tabela_result_metr_time} points out that our model decreased the Inference Time by 75.13\% in \textit{CDPhotoDataset} and by 73.79\% in \textit{DTDDataset}. In addition to the reduction in processing time, Table \ref{tabela_result_storage} shows that OctHU-PageScan uses 93.49\% less disk storage space when compared to the U-Net, having 93.52\% less trainable parameters.

\begin{figure}
    \centering
    \subfigure[baseline]{\includegraphics[width=\columnwidth,keepaspectratio]{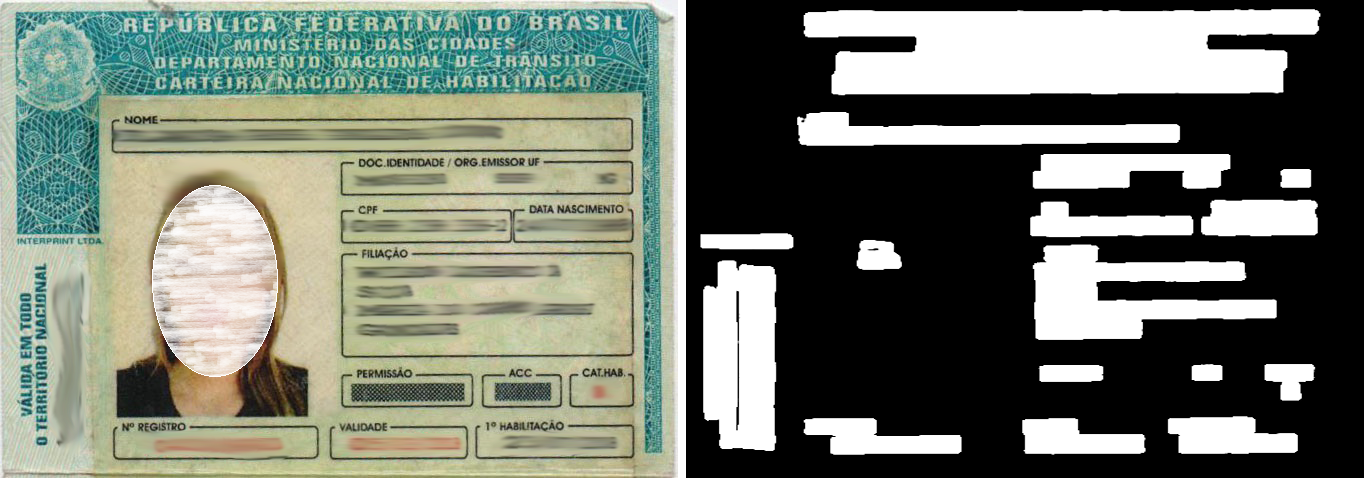}}
    \\
    \subfigure[proposed]{\includegraphics[width=\columnwidth,keepaspectratio]{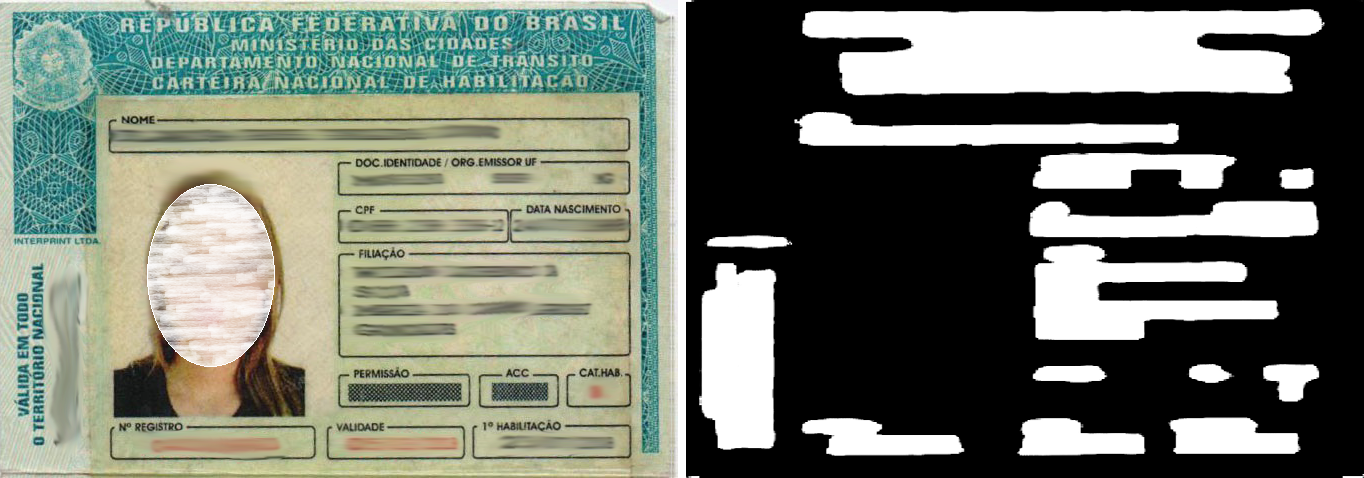}}
    \caption{Qualitative analysis of the result achieved at DTDDataset.}
    \label{RESULT_FIGURE_1}
\end{figure}

Despite the difference in the performance of the methods suggested by the JCS metric, as it is a pixel-based distance metric, this distance may not be significant in qualitative analysis. In Figures \ref{RESULT_FIGURE_1}(a) and \ref{RESULT_FIGURE_1}(b), we show the obtained results on the same image of the \textit{DTDDataset} by both methods. The qualitative analysis suggests the model's performance is very close and possible not impair in the subsequent image recognition tasks while allows a huge decrease in hardware requirements.

\begin{figure}
    \centering
    \subfigure[baseline]{\includegraphics[width=0.49\columnwidth]{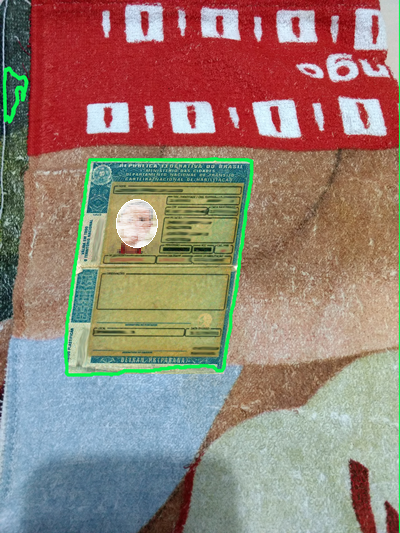}}
    \subfigure[proposed]{\includegraphics[width=0.49\columnwidth]{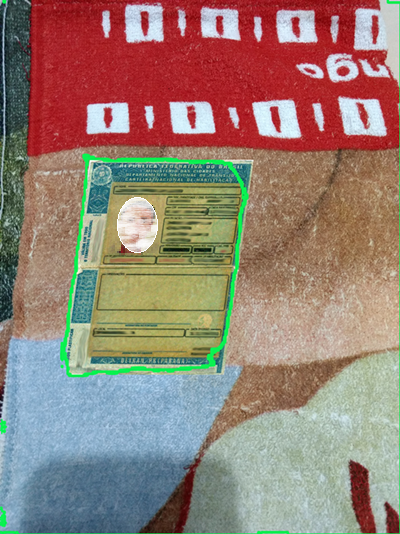}} 
    \caption{Qualitative analysis of the result achieved at CDPhotoDataset.}\label{RESULT_FIGURE_2}
\end{figure}

As in Figure  \ref{RESULT_FIGURE_1}, the qualitative analysis in Figure  \ref{RESULT_FIGURE_2} indicates that although there is a numerical difference in the JCS index, visually, both models behave similarly. Figure \ref{RESULT_FIGURE_2} (a) shows that the U-Net model can identify the edges of the document with excellent precision; however because it has a more significant number of feature maps, it ends up identifying false positives (that is, classifying as document the background pixels). 

In contrast, Figure \ref{RESULT_FIGURE_2} (b) shows that the proposed model, OctHU-Page Scan, detects the edges of the document with less precision in comparison with the standard U-Net. However, this result can be improved by running a low-cost polygon approximation algorithm as a post-processing step.

The reason that the proposed model achieved a slightly lower JCS performance than U-Net is because the U-Net has more feature maps, allowing it to extract more detail from the input images. On the other hand, this more significant number of feature maps present in the U-Net model reflects in a more complex model, with a higher computational cost for training and inference.

Our proposed model, on the other hand, uses octave convolutions, which allows the extraction of high and low-frequency characteristics, allowing the use of a smaller number of features maps and, as a consequence, much fewer parameters and required storage size.

\begin{table}

\renewcommand{\arraystretch}{1.1}

\caption{Comparison of JCS Rate with respective standard deviation between U-Net and proposed model }
\label{tabela_result_metr_jcs}
\centering

\begin{tabularx}{\columnwidth}{YYY}

\toprule
Models & CDPhotoDataset (\%) & DTDDataset (\%) \\

\midrule
U-Net & 0.9916 ($\pm$ 0.0107) & 0.9599 ($\pm$ 0.0144) \\

OctHU-PS & 0.9456 ($\pm$ 0.0619) & 0.7948 ($\pm$ 0.0796) \\

\bottomrule

\end{tabularx}
\end{table}

\begin{table}

\renewcommand{\arraystretch}{1.1}

\caption{Comparative of Inference Time between U-Net and proposed model}
\label{tabela_result_metr_time}
\centering

\begin{tabularx}{\columnwidth}{YYY}

\toprule
Models & CDPhotoDataset (seconds) & DTDDataset (seconds) \\

\midrule
U-Net & 0.0973 & 0.0912 \\

OctHU-PS & \textbf{0.0242} & \textbf{0.0239} \\

\bottomrule

\end{tabularx}
\end{table}

\begin{table}

\renewcommand{\arraystretch}{1.1}

\caption{Comparison in storage space and number of parameters between U-Net and proposed model}
\label{tabela_result_storage}
\centering

\begin{tabularx}{\columnwidth}{YYY}

\toprule
Models & Pre-trained Model Size (.pb file) & Total Trainable params) \\

\midrule
U-Net & 118.3 MB & 30,299,233 \\

OctHU-PS & \textbf{7.7mb} & \textbf{1,963,794} \\

\bottomrule

\end{tabularx}
\end{table}

\section{Conclusion}

The capture and automatic segmentation of personal documents is an essential task to initiate the relationship between companies and users in the online society. As the entities are delegating the capture of these documents to the end-user, they need advanced technology to ensure the quality and feasibility of the captured images. In this context, the image segmentation and the identification of the documents type and text regions are necessary prerequisites for automation.

To address this problem, the present work proposes a Deep Learning model to image segmentation, which is applied for border and text detection in Brazilian ID documents. Segmentation-based applications are essential because it serves as a preprocessing method for OCR algorithms and other image applications. In this sense, we investigated the document image segmentation problem in two levels of granularity. For this, we developed two new different datasets. The first one is the CDPhotoDataset, which has 20,000 images of documents with different levels of noises, rotations, and image resolutions. The second one is DTDDataset, composed of 800 real Brazilian documents and after data augmentation a total of 10,000 images.

The proposed model uses U-Net as inspiration, replacing the Convolutional Layers by Octave Convolutional Layers, therefore, introducing the Fully Octave Convolutional Neural Network. We also reduce the redundancy of feature maps and generate a lighter model. In the proposed architecture, we use encoders and decoders to extract the main features and reconstruct the image with the interest regions highlighted.

Finally, we noticed that while decreasing of 75.13\% and 73.79\% in inference time in CDPhotoDataset and DTDDataset, respectively, the method experimented a decrease in the JCS Rate of 4.6 and 16.51 percentual points in those datasets. Although the JCS rate showed a considerable difference between the models, the qualitative analysis visually shows the results between the models are quite similar. Because of the higher number of feature maps, the standard U-Net has greater accuracy in detection, however, inferring more false positives pixels. While OctHU-PageScan can identify the region satisfactorily, without false positive pixels.

We believe that the results show a possibility of further research on the usage of Octave Convolutions as alternative to standard Convolutions in case of FCNs, in many other image segmentation problems where time and computational resources are critical, expanding the usage of the here named Fully Octave Convolutional Networks (FOCN or FOCNN).

As future work, the authors will investigate other Deep Learning approaches applied for the problem addressed at this work. For improving the JCS Rate, the authors will investigate other powerful Deep Learning models with Quantization \cite{cai2017deep} and Distilling the Knowledge \cite{hinton2015distilling} approaches.

\section{Acknowledgment}

This study was financed in part by: Coordenação de Aperfeiçoamento de Pessoal de Nível Superior - Brasil (CAPES) - Finance Code 001, Fundação de Amparo a Ciência e Tecnologia do Estado de Pernambuco (FACEPE), and Conselho Nacional de Desenvolvimento Científico e Tecnológico (CNPq) - Brazilian research agencies.

\ifCLASSOPTIONcaptionsoff
  \newpage
\fi

\bibliographystyle{ieeetr}
\bibliography{reference.bib}

\begin{thebibliography}{10}

\bibitem{mullins2014know}
R.~R. Mullins, M.~Ahearne, S.~K. Lam, Z.~R. Hall, and J.~P. Boichuk, ``Know
  your customer: How salesperson perceptions of customer relationship quality
  form and influence account profitability,'' {\em Journal of Marketing},
  vol.~78, no.~6, pp.~38--58, 2014.

\bibitem{sharman2008power}
J.~C. Sharman, ``Power and discourse in policy diffusion: Anti-money laundering
  in developing states,'' {\em International Studies Quarterly}, vol.~52,
  no.~3, pp.~635--656, 2008.

\bibitem{mello2012}
C.~A.~B. MELLO, W.~P. SANTOS, and A.~L.~I. OLIVEIRA, {\em Digital Document
  Analysis and Processing}.
\newblock New York: Nova Science Publishers Inc, 2012.

\bibitem{forczmanski2019segmentation}
P.~Forczma{\'n}ski, A.~Smoli{\'n}ski, A.~Nowosielski, and K.~Ma{\l}ecki,
  ``Segmentation of scanned documents using deep-learning approach,'' in {\em
  International Conference on Computer Recognition Systems}, pp.~141--152,
  Springer, 2019.

\bibitem{oliveira2018dhsegment}
S.~A. Oliveira, B.~Seguin, and F.~Kaplan, ``dhsegment: A generic deep-learning
  approach for document segmentation,'' in {\em 2018 16th International
  Conference on Frontiers in Handwriting Recognition (ICFHR)}, pp.~7--12, IEEE,
  2018.

\bibitem{renton2017handwritten}
G.~Renton, C.~Chatelain, S.~Adam, C.~Kermorvant, and T.~Paquet, ``Handwritten
  text line segmentation using fully convolutional network,'' in {\em 2017 14th
  IAPR International Conference on Document Analysis and Recognition (ICDAR)},
  vol.~5, pp.~5--9, IEEE, 2017.

\bibitem{ronneberger2015u}
O.~Ronneberger, P.~Fischer, and T.~Brox, ``U-net: Convolutional networks for
  biomedical image segmentation,'' in {\em International Conference on Medical
  image computing and computer-assisted intervention}, pp.~234--241, Springer,
  2015.

\bibitem{chen2019drop}
Y.~Chen, H.~Fang, B.~Xu, Z.~Yan, Y.~Kalantidis, M.~Rohrbach, S.~Yan, and
  J.~Feng, ``Drop an octave: Reducing spatial redundancy in convolutional
  neural networks with octave convolution,'' {\em arXiv preprint
  arXiv:1904.05049}, 2019.

\bibitem{fan2019accurate}
Z.~Fan, J.~Mo, and B.~Qiu, ``Accurate retinal vessel segmentation via octave
  convolution neural network,'' {\em arXiv preprint arXiv:1906.12193}, 2019.

\bibitem{lundgren2019octshufflemlt}
A.~Lundgren, D.~Castro, E.~B. Lima, and B.~L.~D. Bezerra, ``Octshufflemlt: A
  compact octave based neural network for end-to-end multilingual text
  detection and recognition,'' in {\em 2019 International Conference on
  Document Analysis and Recognition Workshops (ICDARW)}, vol.~4, pp.~37--42,
  IEEE, 2019.

\bibitem{burie2015icdar2015}
J.-C. Burie, J.~Chazalon, M.~Coustaty, S.~Eskenazi, M.~M. Luqman, M.~Mehri,
  N.~Nayef, J.-M. Ogier, S.~Prum, and M.~Rusi{\~n}ol, ``Icdar2015 competition
  on smartphone document capture and ocr (smartdoc),'' in {\em 2015 13th
  International Conference on Document Analysis and Recognition (ICDAR)},
  pp.~1161--1165, IEEE, 2015.

\bibitem{melo2018fully}
V.~K. S.~L. Melo and B.~L.~D. Bezerra, ``A fully convolutional network for
  signature segmentation from document images,'' in {\em 2018 16th
  International Conference on Frontiers in Handwriting Recognition (ICFHR)},
  pp.~540--545, IEEE, 2018.

\bibitem{fierrez2004off}
J.~Fierrez-Aguilar, N.~Alonso-Hermira, G.~Moreno-Marquez, and J.~Ortega-Garcia,
  ``An off-line signature verification system based on fusion of local and
  global information,'' in {\em International workshop on biometric
  authentication}, pp.~295--306, Springer, 2004.

\bibitem{paloma2019}
P.~G.~S. Silva, C.~A. M.~L. Junior, E.~B. Lima, B.~L.~D. Bezerra, and
  C.~Zanchettin, ``Speeding-up the handwritten signature segmentation process
  through an optimized fully convolutional neural network,'' in {\em 15th
  International Conference on Document Analysis and Recognition (ICDAR 2019)},
  vol.~1, IEEE, 2019.

\bibitem{zhang2018shufflenet}
X.~Zhang, X.~Zhou, M.~Lin, and J.~Sun, ``Shufflenet: An extremely efficient
  convolutional neural network for mobile devices,'' in {\em Proceedings of the
  IEEE Conference on Computer Vision and Pattern Recognition}, pp.~6848--6856,
  2018.

\bibitem{kingma2014adam}
D.~P. Kingma and J.~Ba, ``Adam: A method for stochastic optimization,'' {\em
  arXiv preprint arXiv:1412.6980}, 2014.

\bibitem{hamers1989similarity}
L.~Hamers {\em et~al.}, ``Similarity measures in scientometric research: The
  jaccard index versus salton's cosine formula.,'' {\em Information Processing
  and Management}, vol.~25, no.~3, pp.~315--18, 1989.

\bibitem{cai2017deep}
Z.~Cai, X.~He, J.~Sun, and N.~Vasconcelos, ``Deep learning with low precision
  by half-wave gaussian quantization,'' in {\em Proceedings of the IEEE
  Conference on Computer Vision and Pattern Recognition}, pp.~5918--5926, 2017.

\bibitem{hinton2015distilling}
G.~Hinton, O.~Vinyals, and J.~Dean, ``Distilling the knowledge in a neural
  network,'' {\em arXiv preprint arXiv:1503.02531}, 2015.

\end{thebibliography}

\end{document}